\documentclass[runningheads]{llncs}
\usepackage{graphicx}
\usepackage{booktabs}
\usepackage{multirow}
\usepackage{multicol}
\usepackage{amsmath}
\usepackage{amsfonts}
\usepackage{caption}
\usepackage{subcaption}
\usepackage{url}

\newcommand{\bA}{\mathbf{A}}
\newcommand{\bB}{\mathbf{B}}
\newcommand{\bC}{\mathbf{C}}
\newcommand{\bD}{\mathbf{D}}
\newcommand{\bE}{\mathbf{E}}

\newcommand{\bH}{\mathbf{H}}

\newcommand{\bK}{\mathbf{K}}

\newcommand{\bM}{\mathbf{M}}

\newcommand{\bQ}{\mathbf{Q}}

\newcommand{\bS}{\mathbf{S}}

\newcommand{\bV}{\mathbf{V}}
\newcommand{\bW}{\mathbf{W}}

\newcommand{\bb}{\mathbf{b}}

\newcommand{\be}{\mathbf{e}}

\newcommand{\bh}{\mathbf{h}}

\newcommand{\br}{\mathbf{r}}
\newcommand{\bs}{\mathbf{s}}

\newcommand{\bu}{\mathbf{u}}
\newcommand{\bv}{\mathbf{v}}

\newcommand{\ie}{\textit{i.e.}}

\begin{document}
\title{Content Selection Network for Document-grounded Retrieval-based Chatbots}
\titlerunning{Content Selection Network}
\author{Yutao Zhu\inst{1}\thanks{Corresponding author.} \and
Jian-Yun Nie\inst{1} \and
Kun Zhou\inst{2} \and
Pan Du\inst{1} \and
Zhicheng Dou\inst{3}
}
\authorrunning{Y. Zhu et al.}
\institute{Université de Montréal, Québec, Canada\\
% \email{yutao.zhu@umontreal.ca, \{nie,pandu\}@iro.umontreal.ca}\\
\and
School of Information, Renmin University of China\\
% \email{francis\_kun\_zhou@163.com}\\
\and
Gaoling School of Artificial Intelligence, Renmin University of China\\
\email{yutao.zhu@umontreal.ca, \{nie,pandu\}@iro.umontreal.ca,\\francis\_kun\_zhou@163.com, dou@ruc.edu.cn}
}
\maketitle              % typeset the header of the contribution
\begin{abstract}
Grounding human-machine conversation in a document is an effective way to improve the performance of retrieval-based chatbots. However, only a part of the document content may be relevant to help select the appropriate response at a round. It is thus crucial to select the  part of document content relevant to the current conversation context. In this paper, we propose a document content selection network (CSN) to perform explicit selection of relevant document contents, and filter out the irrelevant parts. We show in experiments on two public document-grounded conversation datasets that CSN can effectively help select the relevant document contents to the conversation context, and it produces better results than the state-of-the-art approaches. Our code and datasets are available at \url{https://github.com/DaoD/CSN}.

\end{abstract}
\section{Introduction}
Retrieval-based chatbots such as Microsoft XiaoIce~\cite{DBLP:journals/jzusc/ShumHL18} and Amazon Alexa~\cite{DBLP:journals/corr/abs-1801-03604} are widely used in real-world applications.
% because the selected responses are usually fluent and grammatically correct, and the methods are easy to evaluate.
Given a user input, an upstream retrieval system can provide a set of response candidates, and the retrieval-based chatbot should choose the appropriate one. This mainly relies on a matching score between the context and each candidate response. 
%Recent studies have proposed various neural architectures for the task and achieved great performance~\cite{lowe-etal-2015-ubuntu,wu-etal-2017-sequential,zhou-etal-2018-multi,tao-etal-2019-one,yuan-etal-2019-multi}. %However, in most studies, the response is selected merely based on the context. 
It has been found that the conversation context alone is insufficient in many cases for response selection~\cite{tian-etal-2017-make,zhang-etal-2018-personalizing}. In fact, human conversations are usually also grounded in external knowledge or documents: our responses are strongly related to our knowledge or information contained in the documents at hand. On Reddit, for example, people usually discuss about a document posted at the beginning of a thread, which provides the background topics and basic facts for the following conversations. On Twitter, people may also exchange opinions related to a news article. In these cases, in addition to the conversation context, the document or news article also provides useful background information to guide response selection. A conversation that does not take into account the background information may lead to off-topic responses. This paper deals with the problem of document-grounded conversation - conversation based on a given document ~\cite{zhang-etal-2018-personalizing,arora-etal-2019-knowledge,qin-etal-2019-conversing,DBLP:conf/ijcai/ZhaoTWX0Y19,gu-etal-2019-dually}. 

\begin{figure}[t!]
    \centering
    \includegraphics[width=.9\linewidth]{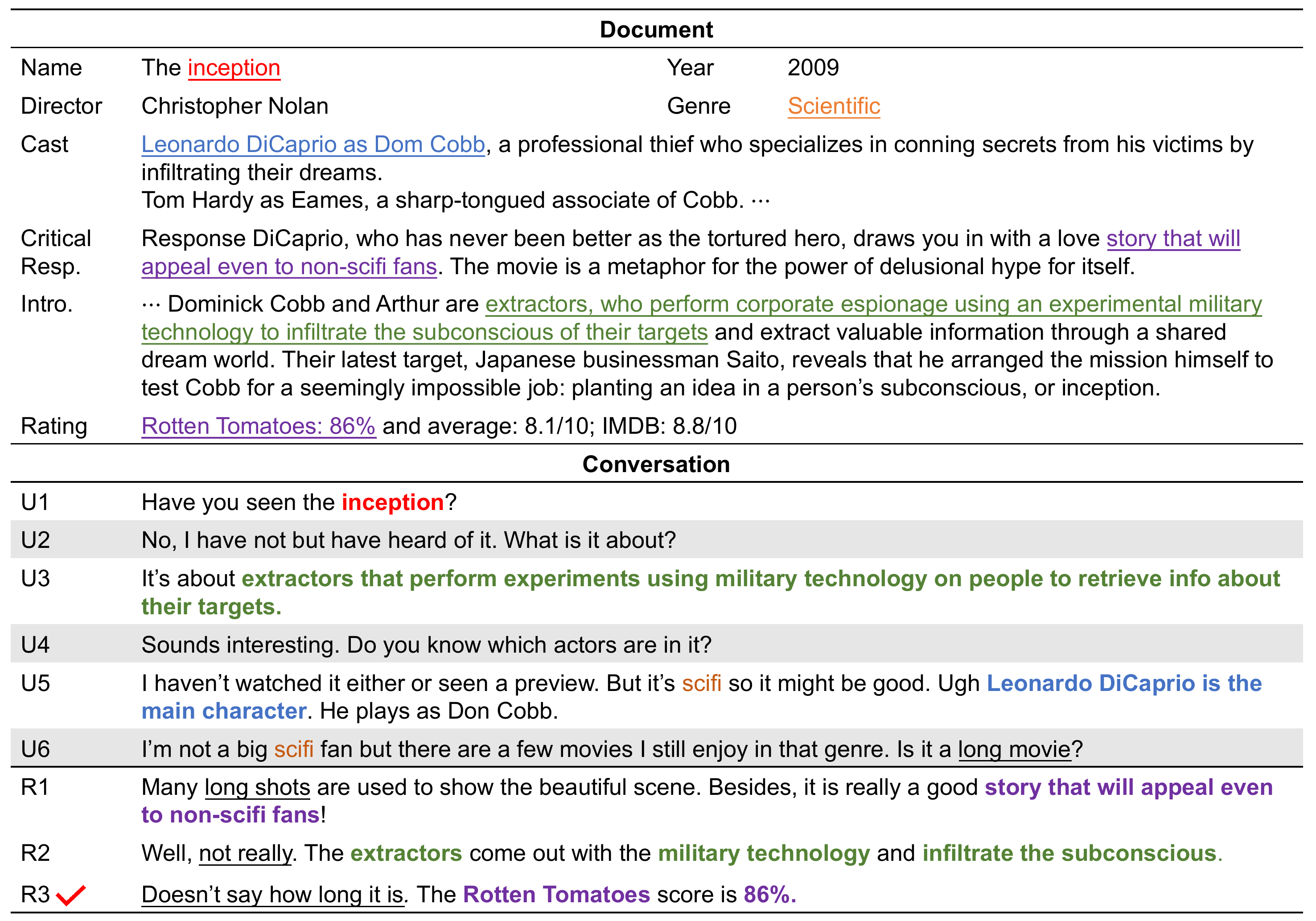}
    \caption{An example in CMUDoG dataset. The words in color correspond to those in the document. R3 is the ground-truth response.}
    \vspace{-10px}
    \label{fig:eg}
\end{figure}

The task of document-grounded response selection is formulated as selecting a good response from a candidate pool that is consistent with the context and relevant to the document. Several existing studies have shown that leveraging the background document can significantly improve  response selection~\cite{zhang-etal-2018-personalizing,DBLP:conf/ijcai/ZhaoTWX0Y19,gu-etal-2019-dually}. Generally, the common strategy is selecting the response based on a combination of context-response matching and document-response matching. The latter can boost the responses that are related to the document content. 
However, a good response does not need to be related to the whole content of the document, but to a small part of it. %the document-response matching should not be estimated using the whole document. It may often be the case that a response relevant to the document is not fully appropriate for the current dialogue context. 
%In fact, when we construct our response based on a document, we may rely on a piece of detailed information from the document. 
The selection of the relevant part of the document is crucial.
%A global document model may fail to capture such detailed information. %We need to exploit the document content at a finer granularity.
%Besides, the document is usually modeled in a general view, thus some detailed information is neglected. This may cause the model select a response inconsistent with the document.

The problem can be illustrated by an example from CMUDoG~\cite{zhou-etal-2018-dataset} in Fig.~\ref{fig:eg}. In this dataset, a movie-related wiki article is used as the grounding document. We can see that the conversation is highly related to the document. R1, R2, and R3 are three candidate responses for U6, and R3 is the desired response.
% Taken in isolation without the document, R1 could be a response to the U6, but it is inconsistent with  the grounding document, which describes a science fiction movie. To determine this inconsistency, we have to retain the fine-grain information about the movie's genre in the representation. %This indicates that the fine-grained information contained in the document should be well-captured.
The wrong response R1 could be highly scored because it shares several key words  with the document (\ie,  document-response matching score is high). However, R1 is not an appropriate response in the current context, which  asks about the length of the movie. This example shows that a correct response is well grounded in the document not because it corresponds to the document content, but because it corresponds to the part relevant to the conversation context. 
%have to consider the conversation context at the same time.  
Therefore, a first challenge is to select the part of the document content relevant to the current conversation context. %choose the most consistent piece of information from the document for response selection. 
R2 looks like a proper response to  U6, yet it conveys similar information as U3, which makes the dialogue less informative. This response could be selected if we use the whole conversation history as conversation context - the response could have a high context-response matching score. In fact, the current context in this example is about the length of the movie. The previous utterances in the history are less relevant. 
This case illustrates the need to well calibrate and model the current conversation context.
% Furthermore, the context-response matching should also be considered in detail, as shallow word matching between ``long movie'' and ``long shots'' may bring noise.
% R2 receives a very high score in existing models~\cite{zhang-etal-2018-personalizing,DBLP:conf/ijcai/ZhaoTWX0Y19,gu-etal-2019-dually} because it shares a lot of words with the document. However, these words from the document are irrelevant to the current conversation step. 
%R3 is the ground-truth response that is both coherent with the conversation context and consistent with some part of the document content.

The two key problems illustrated by the above example (R1 and R2) are not well addressed in previous studies: %(1) Some of the existing studies aggregate the whole document content into a vector representation and then match it with the response~\cite{DBLP:conf/aaai/WuFCABW18,zhang-etal-2018-personalizing,mazare-etal-2018-training}. The global document representation may fail to capture the fine-grain information about move genre. %is process loses a lot of fine-grained document information. 
(1) They usually perform a soft selection of document content by assigning attention weights to them~\cite{DBLP:conf/ijcai/ZhaoTWX0Y19,gu-etal-2019-dually}. Even though the less relevant parts could be assigned lower weights, the cumulative weight of many irrelevant parts could be large, so that they collectively influence the response selection in a wrong direction. 
%different parts of the document are assigned with different weights (\eg, by attention mechanism), the noise information (irrelevant content) may still be retained to some extent, and it may mislead the model to select a wrong response. 
We believe that a key missing element is a proper (hard) selection of the document content that fits the current conversation context, instead of a (soft) weighting. The hard selection of document content is motivated by the following observation: although the whole conversation can cover many aspects described in the grounding document, each of the step is related to only a small part of the document content. For example, in our conversation about a movie, we could discuss about an actor in one step. The selection of such a small part of the content is crucial. This observation advocates a hard selection rather than a soft weighting used in the previous studies.
(2) The existing studies usually use the entire context to determine the weights of parts (sentences) of the document content. This strategy fails to distinguish the current conversation context from the ones in the history. As a result, a past round of conversation could be mistaken as the current one, leading to a redundant response as  illustrated by the R2 example. 
%, there are many chances that the document contents with high weights will be repeatedly utilized during the conversation, leading to redundant responses.

% (1) Most existing approaches focus on information aggregation but ignoring the topic evolution nature of conversations when selecting responses.
In this paper, we propose a \textbf{Content Selection Network} (CSN) to tackle these problems. \textbf{First}, we use a modified \textit{gate mechanism} to implement the document content selection according to the conversation context, before using it to match with the response candidate. The content relevant to the current conversation step will be assigned a higher weight and pass the selection gate, while the irrelevant parts will be blocked. We use the gate mechanism to select sentences or words.
%works at fine-grained levels - on sentences and words, so as to retain fine-grained information. At sentence/word level, each sentence/word in the document is assigned a relevance weight, and only those that have a sufficient weight will be kept.
%while at word level, each word is assigned a relevance weight. 
\textbf{Second}, as the topic usually evolves during the conversation, we determine the current conversation context by focusing on the most recent utterances, rather than on the whole conversation history. To this end, we design a decay mechanism for the history to force the model focusing more on the current dialogue topic. 
% \textbf{Second}, we use a deep matching architecture to model the %matching considered the fine-grained information in the document. The 
% interaction between each selected document sentence and the response. 
%\textbf{Finally}, 
The selected document contents and the conversation context are finally combined to select the candidate response. 

The main contributions of this paper are: (1) We propose a content selection network to explicitly select the relevant sentences/words from the document to complement the conversation context. Our experiments show that this is a much more effective way to leverage the grounding document than a soft weighting. 
%to avoid disorienting conversation by irrelevant document contents. 
(2) We show that document-grounded conversation should focus on the topics in the recent state rather than using the whole conversation context.
On two public datasets for document-grounded conversation, our method outperforms the existing state-of-the-art approaches significantly.
%to select document content.
% exploit document content at fine-grained level rather than taking the entire global content representation.
%consider the document-response matching in a fine-grained view, which help the model better leverage the selected content of the document.

\section{Related Work}
\textbf{Retrieval-based Chatbots}
Existing methods for open-domain dialogue can be categorized into two groups: retrieval-based and generation-based.
Generation-based methods are mainly based on the sequence-to-sequence (Seq2seq) architecture with attention mechanism and aim at generating a new response for conversation context~\cite{shang-etal-2015-neural,DBLP:conf/aaai/SerbanSBCP16,li-etal-2016-diversity,DBLP:conf/aaai/XingWWLHZM17,cai-etal-2019-retrieval}. On the other hand, retrieval-based methods try to find the most reasonable response from a large repository of conversational data~\cite{lowe-etal-2015-ubuntu,wu-etal-2017-sequential,tao-etal-2019-one,yuan-etal-2019-multi}. %Retrieval-based methods are widely used in real conversation systems due to their more fluent and diverse responses and higher efficiency. The typical method is to build a matching model to compute the similarity between the context and the response.
We focus on retrieval-based methods in this paper. 
Early studies use single-turn response selection where the context is a single message~\cite{DBLP:journals/corr/JiLL14,DBLP:conf/nips/HuLLC14}, while recent work considers all previous utterances as context for multi-turn response selection~\cite{wu-etal-2017-sequential,zhou-etal-2018-multi,tao-etal-2019-one,yuan-etal-2019-multi}. In our work, we also consider the whole conversation history (but with decaying weights).

\noindent\textbf{Document-grounded Conversation}
Multiple studies have shown that being grounded in knowledge or document can effectively enhance human-machine conversation~\cite{DBLP:conf/aaai/GhazvininejadBC18,mazare-etal-2018-training,zhang-etal-2018-personalizing,DBLP:conf/ijcai/ZhaoTWX0Y19}. For example, a Seq2seq model is first applied to generate responses based on both conversation history and external knowledge~\cite{DBLP:conf/aaai/GhazvininejadBC18}. An approach using a dually interactive matching network has been proposed, in which 
%Then a persona-based conversation dataset is released. In this dataset, the speakers' personas are provided together with the conversation history. 
context-response matching and document-response matching are performed separately using a shared structure~\cite{gu-etal-2019-dually}. This model achieved state-of-the-art performance
on persona-related conversation~\cite{zhang-etal-2018-personalizing}.
%Zhou et al.~\cite{zhou-etal-2018-dataset} published another dataset in which conversations are grounded in movie-related articles from Wikipedia\footnote{\url{https://www.wikipedia.org}}. 
%In the above methods, the entire document content is taken as a whole to help select responses. As we mentioned earlier, irrelevant document contents could mislead the selection, and the approaches also fail to capture fine-grained information.
Recently, Zhao et al.~\cite{DBLP:conf/ijcai/ZhaoTWX0Y19} proposed a document-grounded matching network that lets the document and the context to attend to each other so as to generate better representations for response selection. Through the attention mechanism, different parts (sentences) of the document  are assigned different weights and will participate in response selection to different extents. However, even though one may expect the noise contents (for the current step)  be assigned with lower weights, they can still participate in response selection. %The situation is the most critical when there is a strong keyword match with the irrelevant parts of the document, as we show by the case of R1 in Fig.~\ref{fig:eg}. 

Our work differs from the existing studies in that we explicitly model the document content selection process and prevent the irrelevant contents from participating in response selection. In addition, we also define the current conversation context by focusing more on recent utterances in the history rather than taking the whole history indistinctly.  These ideas will bring significant improvements compared to the existing methods.
%we pay more attention to the recent topics in the conversation to select document content rather than using the whole conversation context indiscriminately. Our experiment shows that filtering out the irrelevant contents by the recent conversation topics can improve the final response selection performance.

\section{Content Selection Network}
\subsection{Problem Formalization}
Suppose that we have a dataset $\mathcal{D}$, in which each sample is represented as $(c,d,r,y)$, where $c=\{u_{1},\ldots,u_{n}\}$ represents a conversation context with $\{u_{i}\}_{i=1}^{n}$ as utterances; $d=\{s_1,\ldots,s_{m}\}$ represents a document with $\{s_{i}\}_{i=1}^{m}$ as sentences; $r$ is a response candidate; $y \in \{0,1\}$ is a binary label, indicating whether $r$ is a proper response. Our goal is to learn a matching model $g$ from $\mathcal{D}$, such that for a new context-document-response triplet $(c,d,r)$, $g(c,d,r)$ measures the degree of suitability of a response $r$ to the given context $c$ and the document $d$.

% We split this problem into two sub-problems: (1) selecting relevant content from the document; and (2) leveraging the selected content and the context to determine a most reasonable response.

\begin{figure*}[t!]
    \centering
    \includegraphics[width=.9\linewidth]{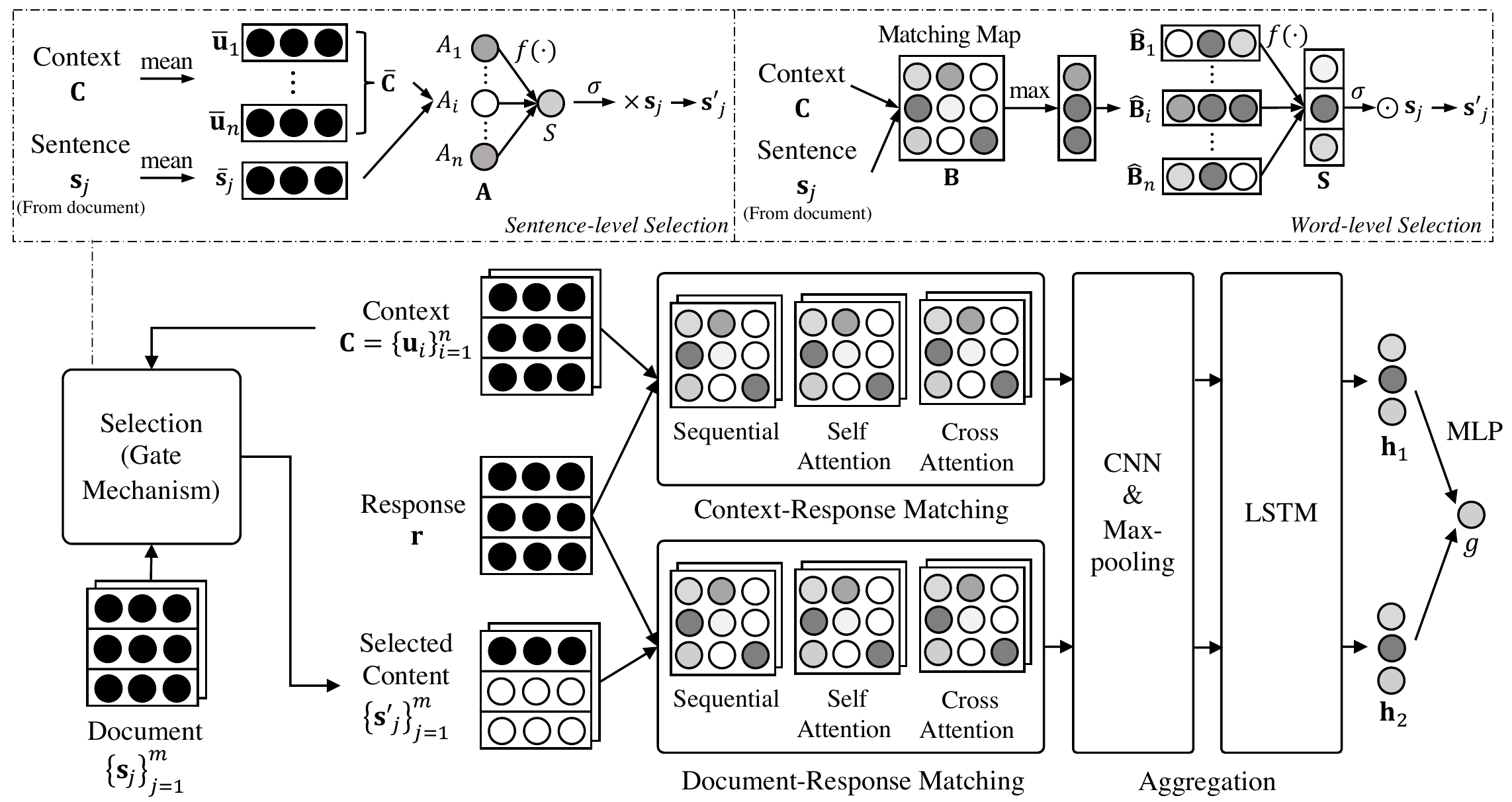}
    \caption{The structure of CSN.}
    \vspace{-10px}
    \label{fig:structure}
\end{figure*}

\subsection{Model Overview}
We propose a content selection network (CSN) to model $g(\cdot,\cdot,\cdot)$, which is shown in Fig.~\ref{fig:structure}. 
Different from the previous work~\cite{wu-etal-2017-sequential,tao-etal-2019-one,gu-etal-2019-dually} which uses the whole document contents, we propose a selection module with a gate mechanism to select the relevant parts of document content based on the context. 
Then, the context-response matching and the document-response matching are modeled based on the sequential, self-attention, and cross-attention representations. Finally, CNNs and RNNs are applied to extract, distill, and aggregate the matching features, based on which the response matching score is calculated. 

\subsection{Representation}
\label{sec:rep}
Consider the $i$-th utterance $u_i=(w^u_1,\cdots,w^u_L)$ in the context, the $j$-th sentence $s_j=(w^s_1,\cdots,w^s_L)$ in the document, and the response $r=(w^r_1,\cdots,w^r_L)$, where $L$ is the number of words\footnote{To simplify the notation, we assume their lengths are the same.}. CSN first uses a pre-trained embedding table to map each word $w$ to a $d_e$-dimension embedding $\be$, \ie, $w \Rightarrow \be$. Thus the utterance $u_i$, the sentence $s_j$, and the response $r$ are represented by matrices $\bE^{u_i}=(\be^{u_i}_1,\cdots,\be^{u_i}_L)$, $\bE^{s_j}=(\be^{s_j}_1,\cdots,\be^{s_j}_L)$, and $\bE^{r}=(\be^{r}_1,\cdots,\be^{r}_L)$, respectively.
Then, CSN encodes the utterances, sentences and responses by bi-directional long short-term memories (BiLSTM)~\cite{DBLP:journals/neco/HochreiterS97} to obtain their sequential representations: $\bu_{i} = \text{BiLSTM}(\bE^{u_i})$, $\bs_{j} = \text{BiLSTM}(\bE^{s_j}) $, $\br = \text{BiLSTM}(\bE^r)$. 
Note that the parameters of these BiLSTMs are shared in our implementation.
% and all these representations have a dimension of $L\times 2d$, where $d$ is the size of the hidden states. 
The whole context is thus represented as $\bC=[\bu_1,\cdots,\bu_n]$.
With the BiLSTM, the sequential relationship and dependency among words in both directions are expected to be encoded into hidden vectors. 
 
\subsection{Content Selection}
In document-grounded conversation, the document usually contains a large amount of diverse information, but only a part of it is related to the current step of the conversation. %Existing methods merely focus on how to better match the document with the response candidate. Instead, we deal with the important problem of document content selection given the conversation context.
%leading to unavoidable noise in the matching phase. 
To select the relevant part of document contents, we propose a content selection phase by a \textit{gate mechanism}, which is based on the relevance between the document and the context. We design the gate mechanism at two different levels, \ie, sentence-level and word-level, to capture relevant information at different granularities. If the sentences/words in the document are irrelevant to the current conversation, they will be filtered out. This is an important difference from the traditional gating mechanism, in which elements are assigned different attention weights, but no element is filtered out.
We use the conversation context to control the gate, which contains several previous turns of conversation. Along the turns, the conversation topic gradually changes. The most important topic is that of the most recent turn, while more distant turns are less important. To reflect this fact, we design a decay mechanism on the history to assign a higher importance to the recent context than to the more distant ones.
The selection process is automatically trained with the whole model in an end-to-end manner.

\noindent\textbf{Sentence-level Selection}
\label{sec:sent}
Let us first explain how document sentences are selected according to conversation context.
%As a document contains a number of sentences, we first consider selecting sentences according to their relevance with the context. 
Considering the context $c=(u_1,\cdots,u_n)$ and the $j$-th sentence $s_j$ in the document, CSN computes a score for the sentence $s_j$ by measuring its matching degree with the current dialogue context. 
In particular, CSN first obtains the sentence representations of the context $c$ and the sentence $s_j$ by mean-pooling over the word dimension of their sequential representations:
\begin{align}
    \bar{\bC} = \underset{dim=2}{\text{mean}}(\bC), \quad \bar{\bs}_j = \underset{dim=1}{\text{mean}}(\bs_j),
\end{align}
where $\bar{\bC}\in\mathbb{R}^{n\times 2d}$ and $\bar{\bs}_j\in\mathbb{R}^{2d}$.
Then CSN computes a sentence-level matching vector $\bA$ by cosine similarities:
\begin{align}
    \bA = \cos(\bar{\bC},\bar{\bs}_j).
\end{align}
We can treat $\bA\in\mathbb{R}^n$ as a similarity array $\bA=[A_1,\cdots,A_n]$ and compute a matching score $S$ for the sentence $s_j$ by fusing the similarity scores:
\begin{align}
    S = f(A_1,A_2,\cdots,A_n). \label{eq:f2}
\end{align}
The fusion function $f(\cdot)$ can be designed in different ways, which will be discussed later.
After obtaining the matching scores for sentences, we select the relevant sentences and update their representations as follows:
\begin{align}
    S' = S \times (\sigma (S) \geq \gamma), \quad
    \bs_j' = S' \times \bs_j, \label{eq:up1}
\end{align}
where $\sigma(\cdot)$ is the Sigmoid function and $\gamma$ is a hyperparameter of the gate threshold. By this means, we will filter out a sentence $s_j$ if its relevance score is below $\gamma$. The filtering is intended to remove the impact of clearly irrelevant parts of document content. %On the contrary, the relevant sentences will be assigned higher weights and contribute more in the matching phase.

\noindent\textbf{Word-level Selection}
\label{sec:word}
In the sentence-level selection, all words in a sentence are assigned  the same weights. We can further perform a selection of words by computing a score for each word in the sentence.
Specifically, CSN constructs a word-level matching map through the attention mechanism as follows:
\begin{align}
    \bB = \bv^{\top} \tanh (\bs_j^{\top}\bW_1\bC + \bb_1),
\end{align}
where $\bW_1\in\mathbb{R}^{2d \times 2d \times h}$, $\bb_1\in\mathbb{R}^{h}$ and $\bv\in\mathbb{R}^{h\times 1}$ are parameters. $\bB\in\mathbb{R}^{n\times L\times L}$ is the word-alignment matrix between the context and the document sentence. Then, to obtain the most important matching features between $s_j$ and each utterance in the context, CSN conducts a max-pooling operation as follows:
\begin{align}
    \hat{\bB} = \max_{dim=3} \bB,
\end{align}
where $\hat{\bB}\in\mathbb{R}^{n\times L}$, and it can be represented in an array form as $\hat{\bB} = [\hat{\bB}_1,\cdots,\hat{\bB}_n]$. The element $\hat{\bB}_i\in\mathbb{R}^{L}$ contains $L$ local matching signals for all words in the document sentence $s_j$ with respect to the utterance $u_i$. Thereafter, CSN applies a fusion function to combine these local matching signals and obtains a global matching vector:
\begin{align}
    {\bS} = f(\hat{\bB}_1, \hat{\bB}_2, \cdots, \hat{\bB}_n). \label{eq:f1}
\end{align}
$\bS\in\mathbb{R}^{L}$ thus contains $L$ global matching scores for all words in $\bs_j$ to the whole context. 
In the next step, %After obtaining the matching scores, 
CSN selects the relevant words in the document and updates the document representation as follows:
\begin{align}
    \bS' = \bS \odot (\sigma (\bS) \geq \gamma), \quad \bs_j' = \bS' \odot \bs_j, \label{eq:up2}
\end{align}
where $\odot$ is the element-wise product. Different from the sentence-level matching score $S'$ in Equation~\ref{eq:up1}, the word-level matching score $\bS'$ is a vector containing weights for different words.

\noindent\textbf{Fusion Function}
\label{sec:ff}
The fusion function $f(\cdot)$ in Equation (\ref{eq:f2}) and (\ref{eq:f1}) is used to aggregate the matching signals with each utterance in the context. Our fusion strategies attribute different weights to the utterances in the conversation history.
Two different functions are considered: (1) Linear combination -- the weight of each matching signal is learned during the model training. Ideally, an utterance containing more information about the conversation topic will contribute more to the selection of document content. (2) Linear combination with decay factors. This method assumes that the topic gradually changes along the conversation and the response is usually highly related to the most recent topic in the context. Therefore, we use a decay factor $\eta\in[0,1]$ on the utterances in the context to decrease their importance when they are far away. The matching scores are then computed as:
\begin{align}
    A_i &= A_i * \eta^{n - i}, \quad (\text{sentence-level}) &
    \hat{\bB}_i &= \hat{\bB}_i * \eta^{n - i}. \quad (\text{word-level})
\end{align}
The decay factor $\eta$ is a hyperparameter. 
Note that when $\eta=1$, it degenerates to the normal linear combination. 
%CSN is equipped with the latter fusion function by default, and we test both fusion functions in our experiments.

\subsection{Matching and Aggregation}
The next problem is to select the appropriate response by leveraging the selected document parts. Following a recent study~\cite{gu-etal-2019-dually}, CSN uses a dually interactive matching structure (as shown in Fig.~\ref{fig:structure}) to determine  context-response matching and document-response matching, where the two kinds of matching features are modeled by the same structure.

Based on the recent work~\cite{wu-etal-2017-sequential,zhou-etal-2018-multi,yuan-etal-2019-multi} that constructs different matching feature maps, in addition to using the sequential representations of the sentences, CSN also uses matching on both self-attention and cross-attention representations. Given the sequential representations of the context $\bC=[\bu_{1},\cdots,\bu_{n}]$, the document $\bD=[\bs'_1,\cdots,\bs'_m]$, and the response candidate $\br$, CSN first constructs a word-word similarity matrix $\bM_1$ by dot product and cosine similarity:
\begin{align}
    \mathbf{M}^{cr}_1 &= \bC\bH_1\br^{\top} \oplus \cos(\bC,\br), & \mathbf{M}^{dr}_1 &= \bD\bH_1\br^{\top} \oplus \cos(\bD,\br),
\end{align}
where $\bH_1\in \mathbb{R}^{2d\times 2d}$ is a parameter, and $\oplus$ is the concatenation operation. 

To better handle the gap in words between two word sequences, CSN applies the attentive module, which is similar to that used in Transformer~\cite{DBLP:conf/nips/VaswaniSPUJGKP17}. The input of an attentive module consists of three sequences, namely query ($\bQ$), key ($\bK$), and value ($\bV$). The output is a new representation of the query and is denoted as $f_\text{ATT}(\bQ,\bK,\bV)$ in the remaining description. 

At first, CSN uses the attentive module over the word dimension to construct multi-grained representations, which is formulated as:
\begin{align}
    \hat{\bC} &= f_\text{ATT}(\bC,\bC,\bC), &
    \hat{\bD} &= f_\text{ATT}(\bD,\bD,\bD), &
    \hat{\br} &= f_\text{ATT}(\br,\br,\br).
\end{align}
The second similarity matrix $\bM_2$ is computed based on these self-attention representations:
\begin{align}
    \bM^{cr}_2 &= \hat{\bC}\bH_2\hat{\br}^{\top} \oplus \text{cos}(\hat{\bC},\hat{\br}), &
    \bM^{dr}_2 &= \hat{\bD}\bH_2\hat{\br}^{\top} \oplus \text{cos}(\hat{\bD},\hat{\br}).
\end{align}
Then, another group of attentive modules (cross-attention) is also applied to represent semantic dependency between the context, the document, and the response candidate:
\begin{align}
    \tilde{\bC} &= f_\text{ATT}(\bC,\br,\br), & \tilde{\br}^{c} &= f_\text{ATT}(\br,\bC,\bC),\\
    \tilde{\bD} &= f_\text{ATT}(\bD,\br,\br), & \tilde{\br}^{d} &= f_\text{ATT}(\br,\bD,\bD).
\end{align}
Next, CSN also constructs a similarity matrix $\bM_3$ as:
\begin{align}
    \bM^{cr}_3 &= \tilde{\bC}\bH_3\tilde{\br}^{c\top} \oplus \text{cos}(\tilde{\bC},\tilde{\br}^{c}), &
    \bM^{dr}_3 &= \tilde{\bD}\bH_3\tilde{\br}^{d\top} \oplus \text{cos}(\tilde{\bD},\tilde{\br}^{d}).
\end{align}

The above matching matrices are concatenated  into two matching cubes:
\begin{align}
    \bM^{cr}&=\bM^{cr}_1\oplus\bM^{cr}_2\oplus\bM^{cr}_3, &
    \bM^{dr}&=\bM^{dr}_1\oplus\bM^{dr}_2\oplus\bM^{dr}_3.
\end{align}
Then CSN applies a CNN with max-pooling operation to extract matching features from $\bM^{cr}$ and $\bM^{dr}$. The output feature maps are flattened as matching vectors. As a result, we obtain two series of matching vectors: (1) between the context and the response $\bv^{cr}=[\bv^{u_1},\cdots,\bv^{u_n}]$; and (2) between the selected document and the response $\bv^{dr}=[\bv^{s_1},\cdots,\bv^{s_m}]$. 

Finally, CSN applies LSTMs to aggregate these two series of matching vectors into two hidden vectors (the last hidden states of the LSTMs):
\begin{align}
    \bh_1 = \text{LSTM}(\bv^{cr}), \quad \bh_2 = \text{LSTM}(\bv^{dr}).    
\end{align}
These vectors are concatenated together and used to compute the final matching score by an MLP with a Sigmoid activation function:
\begin{align}
    g(c,d,r) = \sigma\big({\rm MLP}(\bh_1 \oplus \bh_2)\big).
\end{align}

CSN learns $g(c,d,r)$ by minimizing the following cross-entropy loss with $\mathcal{D}$:
\begin{align}
     \mathcal{L}(\theta) = -\sum_{(y, c, d, r) \in \mathcal{D}}[y\log({g(c,d,r)}) + (1-y)\log({1-g(c,d,r)})].
\end{align}

\section{Experiments}

\subsection{Dataset}
We conduct experiments on two public datasets. 

\noindent\textbf{PersonaChat}~\cite{zhang-etal-2018-personalizing} contains multi-turn dialogues with user profiles. The goal is to generate/retrieve a response that corresponds to the user profile, which is used as a grounding document~\cite{zhang-etal-2018-personalizing}. 
This dataset consists of 8,939 complete dialogues for training, 1,000 for validation, and 968 for testing. 
Response selection is conducted at every turn of a dialogue, and the ratio of the positive and the negative samples is 1:19 in training, validation, and testing sets, resulting in 1,314,380 samples for training, 156,020 for validation, and 150,240 for testing. 
Positive responses are real human responses while negative ones are randomly sampled from other dialogues. 
% There are 955 possible personas for training, 100 for validation, and 100 for testing, each consisting of at least 3 profile sentences.
To prevent the model from taking advantage of trivial word overlap, the revised version of the dataset modified the persona profiles by rephrasing, generalizing, or specializing sentences, making the task much more challenging. We use ``revised'' and ``original'' to indicate the different versions of the dataset. 

\noindent\textbf{CMUDoG}~\cite{zhou-etal-2018-dataset} is designed specifically for  document-grounded conversation. During the conversation, the speakers are provided with a movie-related wiki article. Two scenarios are considered: (1) Only one speaker has access to the article thus she should introduce the movie to the other; (2) Both speakers have access to the article thus they have a discussion. We use the dataset provided by~\cite{DBLP:conf/ijcai/ZhaoTWX0Y19}, where the data of both scenarios are merged because the size of each dataset is relatively small. Notice that the model is only asked to select a response for the user who has access to the document. The ratio of the positive and the negative is 1:19 in training, validation, and testing sets.
This results in 723,180 samples for training, 48,500 for validation, and 132,740 for testing. 

Following previous work~\cite{DBLP:conf/ijcai/ZhaoTWX0Y19}, we employ recall at position $k$ as evaluation metrics (R$@k$), where $k=\{1,2,5\}$. For a single sample, if the only positive candidate is ranked within top $k$ positions, then R$@k = 1$, otherwise, R$@k = 0$. The final value  is the average over all test samples. Note that $R@1$ is equivalent to hits$@1$ that is used in related work~\cite{zhang-etal-2018-personalizing,gu-etal-2019-dually}.

\begin{table*}[t!]
    \centering
    \small
    \caption{Experimental results on all datasets.}
    \begin{tabular}{lccccccccc}
    \toprule
        % & \multicolumn{6}{c|}{PersonaChat} & \multicolumn{3}{c|}{CMUDoG} \\
    % \midrule
        & \multicolumn{3}{c}{PersonaChat-Original} & \multicolumn{3}{c}{PersonaChat-Revised} & \multicolumn{3}{c}{CMUDoG} \\
    \cmidrule(lr){2-4} \cmidrule(lr){5-7} \cmidrule(lr){8-10}
        & \textbf{R@1} & \textbf{R@2} & \textbf{R@5} & \textbf{R@1} & \textbf{R@2} & \textbf{R@5} & \textbf{R@1} & \textbf{R@2} & \textbf{R@5}  \\
    \midrule
        Starspace & 49.1 & 60.2 & 76.5 & 32.2 & 48.3 & 66.7 & 50.7 & 64.5 & 80.3 \\
        Profile & 50.9 & 60.7 & 75.7 & 35.4 & 48.3 & 67.5 & 51.6 & 65.8 & 81.4 \\
        KV Profile & 51.1 & 61.8 & 77.4 & 35.1 & 45.7 & 66.3 & 56.1 & 69.9 & 82.4 \\
        Transformer & 54.2 & 68.3 & 83.8 & 42.1 & 56.5 & 75.0 & 60.3 & 74.4 & 87.4 \\
        DGMN & 67.6 & 81.3 & 93.3 & 56.7 & 73.0 & 89.0 & 65.6 & 78.3 & 91.2 \\
        DIM & 75.5 & 87.5 & 96.5 & 68.3 & 82.7 & 94.4 & 59.6 & 74.4 & 89.6 \\
    \midrule
        CSN-sent & 77.5 & 88.8 & 96.8 & 70.1 & 83.4 & 95.1 & \textbf{70.1} & 82.5 & \textbf{94.3} \\
        CSN-word & \textbf{78.1} & \textbf{89.0} & \textbf{97.1} & \textbf{71.3} & \textbf{84.2} & \textbf{95.5} & 69.8 & \textbf{82.7} & 94.0 \\
    \bottomrule
    \end{tabular}
    % }
    \vspace{-10px}
    \label{tab:result}
\end{table*}

\subsection{Baseline Models}
We compare CSN using sentence-level and word-level selection (denoted as CSN-sent and CSN-word respectively) with the following models:

(1) Starspace~\cite{DBLP:conf/aaai/WuFCABW18} concatenates the document with the context as a long sentence and learns its similarity with the response candidate by optimizing the embeddings using the margin ranking loss and $k$-negative sampling. Matching is done by cosine similarity of the sum of word embeddings.

(2) Profile Memory Network~\cite{zhang-etal-2018-personalizing} uses a memory network with the context as input, then performs attention over the document to find relevant sentences. The combined representation is used to select the response. This model relies on the attention mechanism to weigh document contents.

(3) Key-value (KV) Profile Memory Network~\cite{zhang-etal-2018-personalizing} uses dialogue histories as keys and the next dialogue utterances as values. In addition to the memory of the document, this model has a memory of past dialogues that can  influence the response selection.

(4) Transformer~\cite{DBLP:conf/nips/VaswaniSPUJGKP17} is used in \cite{mazare-etal-2018-training} as an encoder for the context, document, and response. The obtained representations are input to a memory network to conduct matching in the same way as in Profile Memory Network. 

(5) DGMN~\cite{DBLP:conf/ijcai/ZhaoTWX0Y19} is the state-of-the-art model on the CMUDoG dataset. It employs a cross attention mechanism between the context and document and obtains a context-aware document representation and a document-aware context representation. The two representations and the original context representation are all matched with the response representation. The three matching features are finally combined to output the matching score.

(6) DIM~\cite{gu-etal-2019-dually} is the state-of-the-art model on the PersonaChat dataset. It applies a dually interactive matching structure to model the context-response matching and document-response matching respectively. DIM conducts representation, matching, and aggregation by multiple BiLSTMs, and the final matching features are used to compute the matching score by an MLP.

\subsection{Implementation Details}
We use PyTorch~\cite{DBLP:conf/nips/PaszkeGMLBCKLGA19} to implement the model. A 300-dimensional GloVe embedding~\cite{pennington-etal-2014-glove} is used on all datasets. On PersonaChat, another 100-dimensional Word2Vec~\cite{DBLP:conf/nips/MikolovSCCD13} embedding provided by~\cite{gu-etal-2019-dually} is used. Dropout~\cite{DBLP:journals/jmlr/SrivastavaHKSS14} with a rate of 0.2 is applied to the word embeddings. All hidden sizes of the RNNs are set as 300.
Two convolutional layers have 32 and 64 filters with the kernel sizes as [3, 3] and [2, 2].
AdamW~\cite{DBLP:conf/iclr/LoshchilovH19} is employed for optimization with a batch size of 100. The initial learning rate is 0.001 and is decayed by 0.5 when the performance on the validation set is not increasing. 
% The details of the preprocessing for the two datasets are provided in Appendix~\ref{sec:data}.

\subsection{Experimental Results}
The experimental results are shown in Table~\ref{tab:result}. The results on all three datasets indicate that our CSN outperforms all baselines, including DGMN and DIM, which are two state-of-the-art models. On the PersonaChat dataset, both CSN-word and CSN-sent achieve statistically significant improvements ($p$-value $\leq$ 0.05) compared with DIM, which is the best model on this dataset. 
In general, CSN-word performs better than CSN-sent, indicating the word-level selection is more able to select fine-grained document contents than the sentence-level selection. This comparison also confirms our intuition that it is advantageous for document-grounded conversation to rely on fine-grained information from the document.
%to locate the useful content from document. Indeed, there are some overlapping words or synonyms between the context and the document, so it may be easier for CSN-word to judge if a word in the document is related to the current conversation. 
On CMUDoG, the two document content selection strategies work equally well. We explain this by the fact that the grounding document is longer in this dataset, and there is no obvious reason that one level of selection can determine more relevant parts than another. Nevertheless, both selection strategies show clear advantages over the baseline methods without selection.
%the number of words in the context is far less than that in the document in this dataset, so there is less clear advantage for word-level selection.

Compared with other baselines that represent the whole document as a single vector, DGMN, DIM and our CSN consider fine-grained matching between parts of the document and response. We can see that these models achieve clearly better performances, confirming the necessity to use parts of the document rather than the whole document.
%On the other hand, these models use the conversation context to select or weigh sentences in the document to match with the response. 
% We see that these models achieve clearly better performances.
% thus the performance is better. 
%This shows the benefit of selecting relevant sentences from the document. 
However, DGMN and DIM only assign attention weights to sentences according to the context, without eliminating low-weighted ones. %As a result, irrelevant sentences are still involved in response selection, albeit with small weights. 
In contrast, our CSN model %explicitly selects only a few sentences or words that are strongly relevant to the context, and 
filters out all the irrelevant parts. In so doing, we expect the model not to be influenced by clearly irrelevant parts. 
%the conversation can leverage the strongly relevant part of the document rather than be blurred by the entire document contents. 
As we can see in the experimental results, CSN achieves significantly higher performance than DGMN and DIM on all the datasets, confirming the usefulness of explicit selection (and filtering) of document contents.

\begin{figure}[t!]
    \centering
    \begin{subfigure}[b]{0.48\linewidth}
        \centering
        \includegraphics[width=.8\textwidth]{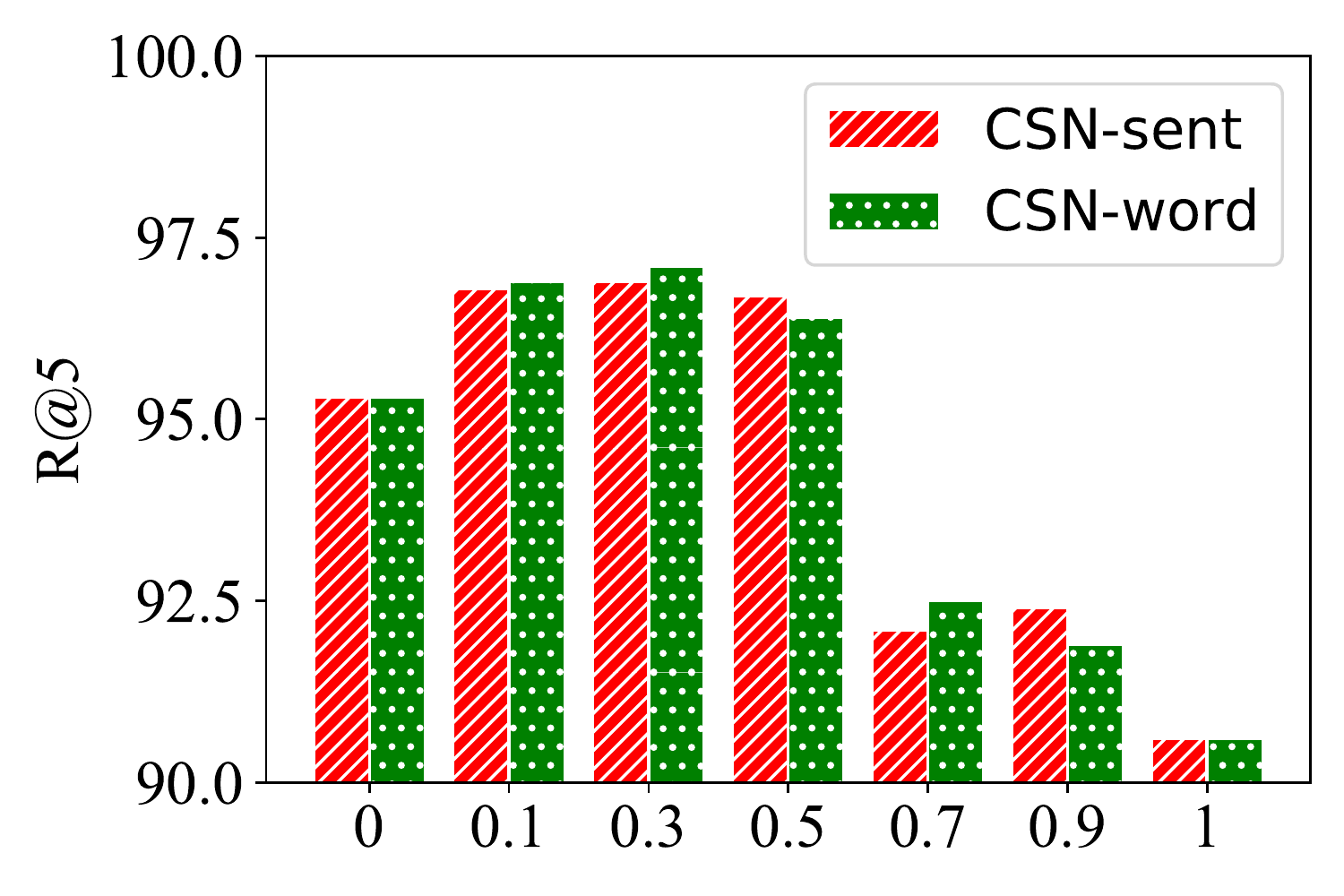}
        \caption{Effect of $\gamma$.}
        \label{fig:gamma}
    \end{subfigure}
    \hfill
    \begin{subfigure}[b]{0.48\linewidth}
        \centering
        \includegraphics[width=.8\textwidth]{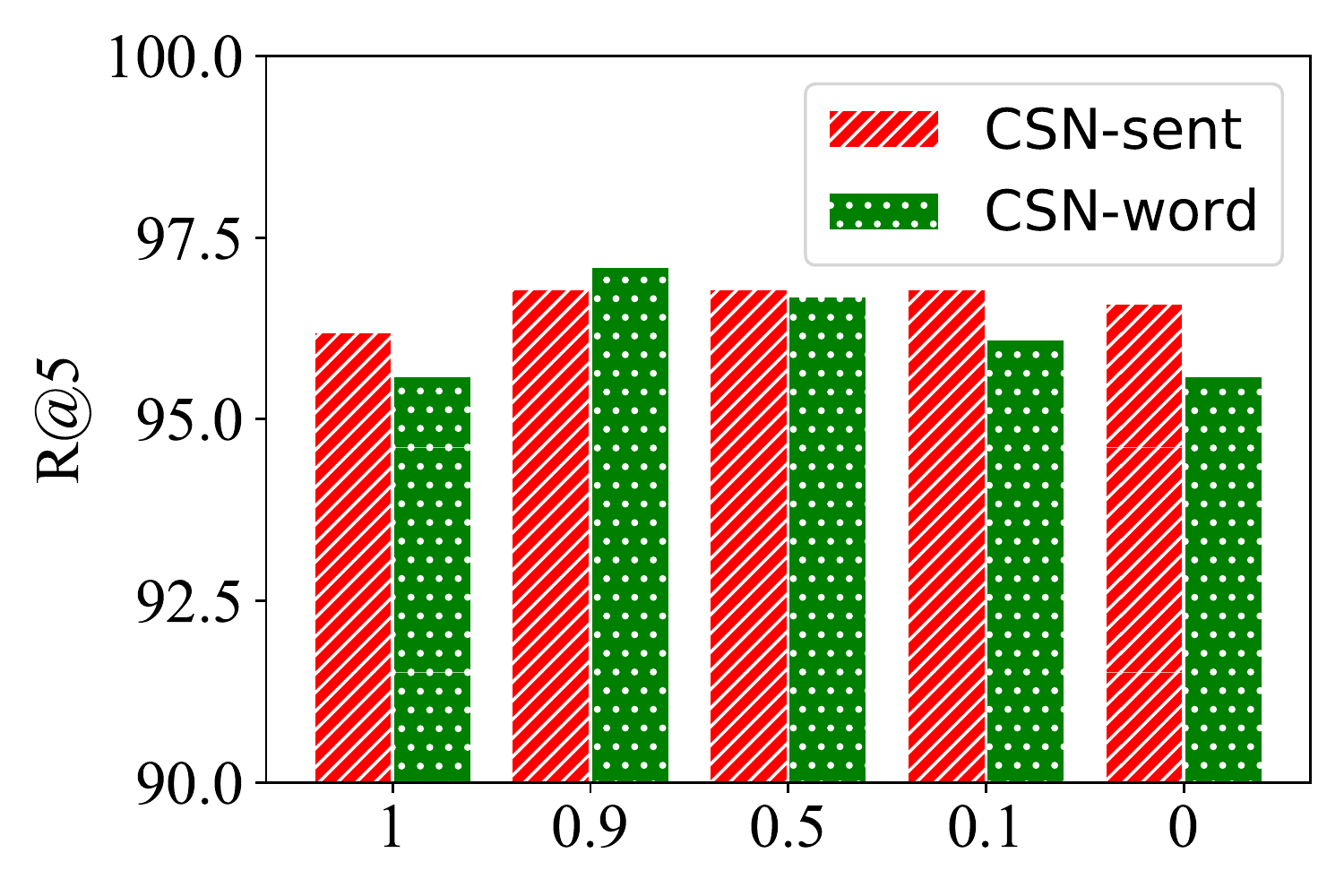}
        \caption{Effect of $\eta$.}
        \label{fig:eta}
    \end{subfigure}
    \caption{Performance of different $\gamma$ and $\eta$ settings on original PersonaChat.}
    \vspace{-10px}
    \label{fig:hp}
\end{figure}
\noindent\textbf{Effect of Content Selection}
The hyperparameter $\gamma$ in Equation (\ref{eq:up1}) and (\ref{eq:up2}) controls how much the document content is selected.
We test the effect of this hyperparameter on the original PersonaChat dataset. Fig.~\ref{fig:gamma} shows that if  $\gamma$ is too small or too large, too much or too little information from the document may be selected. 
%In both cases, the document is not well leveraged. 
In particular,  when $\gamma=0$ -- the whole document content is kept, the performance drops a lot. This strategy is comparable to that used in the existing models DIM and DGMN based on attention. We see again the usefulness of explicit document content filtering. On the other hand, when  $\gamma=1$, \ie, no document content is selected, it degenerates to non document-grounded response selection and the performance also drops sharply. %This shows the great impact of grounding document in conversation. 
The best setting of $\gamma$ is around 0.3 for both CSN-sent and CSN-word, which retains an appropriate amount of relevant document content for response matching.

\noindent\textbf{Effect of Decaying Factor}
The decay factor $\eta$ works as prior knowledge to guide the model focusing more on the recent utterances. A lower $\eta$ means the previous utterances have less contribution in the selection of the document. ``$\eta=1$'' corresponds to the model with a normal linear combination (the first kind of fusion function). Based on the results, we can see that our decaying strategy ($\eta=0.9$)  performs the best. This confirms our assumption that focusing more on the recent topic of the conversation is helpful. However, when $\eta=0$, only the last utterance in the history is used and the performance is lower. This illustrates the necessity of using a larger context. 

\section{Conclusion and Future Work}
In this paper, we proposed a document content selection network to select the relevant content to ground the conversation. We designed a gate mechanism that uses conversation context to retain the relevant document contents while filtering out irrelevant parts. In addition, we also use a decay factor on the conversation history to focus on more recent utterances. 
Our experiments on two large-scale datasets for document-grounded response selection demonstrated the effectiveness of our model. We showed that both document content selection (and filtering) and the use of decay factor contributed in increasing the effectiveness of response selection.
%to be more effective, document-grounded conversation should select only the relevant contents to ground the current conversation context. 
As a future work, %In this work, content selection is done at sentence or word level.
it would be interesting to study if the selection can be done at topic level, in addition to sentence and word levels.


% \bibliography{mybib}
%
\end{document}